\pdfoutput=1

\documentclass[11pt]{article}

\usepackage{ACL2023}

\usepackage{times}
\usepackage{latexsym}

\usepackage{amsmath}
\usepackage{multirow}
\usepackage{graphicx}
\usepackage{stfloats}

\usepackage[T1]{fontenc}

\usepackage[utf8]{inputenc}

\usepackage{microtype}

\usepackage{inconsolata}

\title{Leveraging Explicit Procedural Instructions for Data-Efficient Action Prediction}

\author{Julia White \and Arushi Raghuvanshi \and Yada Pruksachatkun \\ Infinitus Systems, Inc. \\ \texttt{\{julia.white,arushi,yada.pruksachatkun\}@infinitus.ai}}

\begin{document}
\maketitle
\begin{abstract}
Task-oriented dialogues often require agents to enact complex, multi-step procedures in order to meet user requests. While large language models have found success automating these dialogues in constrained environments, their widespread deployment is limited by the substantial quantities of task-specific data required for training. The following paper presents a data-efficient solution to constructing dialogue systems, leveraging explicit instructions derived from agent guidelines, such as company policies or customer service manuals. Our proposed Knowledge-Augmented Dialogue System (KADS) combines a large language model with a knowledge retrieval module that pulls documents outlining relevant procedures from a predefined set of policies, given a user-agent interaction. To train this system, we introduce a semi-supervised pre-training scheme that employs dialogue-document matching and action-oriented masked language modeling with partial parameter freezing. We evaluate the effectiveness of our approach on prominent task-oriented dialogue datasets, Action-Based Conversations Dataset and Schema-Guided Dialogue, for two dialogue tasks: action state tracking and workflow discovery. Our results demonstrate that procedural knowledge augmentation improves accuracy predicting in- and out-of-distribution actions while preserving high performance in settings with low or sparse data.
\end{abstract}

\begin{figure}[t]
\begin{center}
\includegraphics[width=7.5cm]{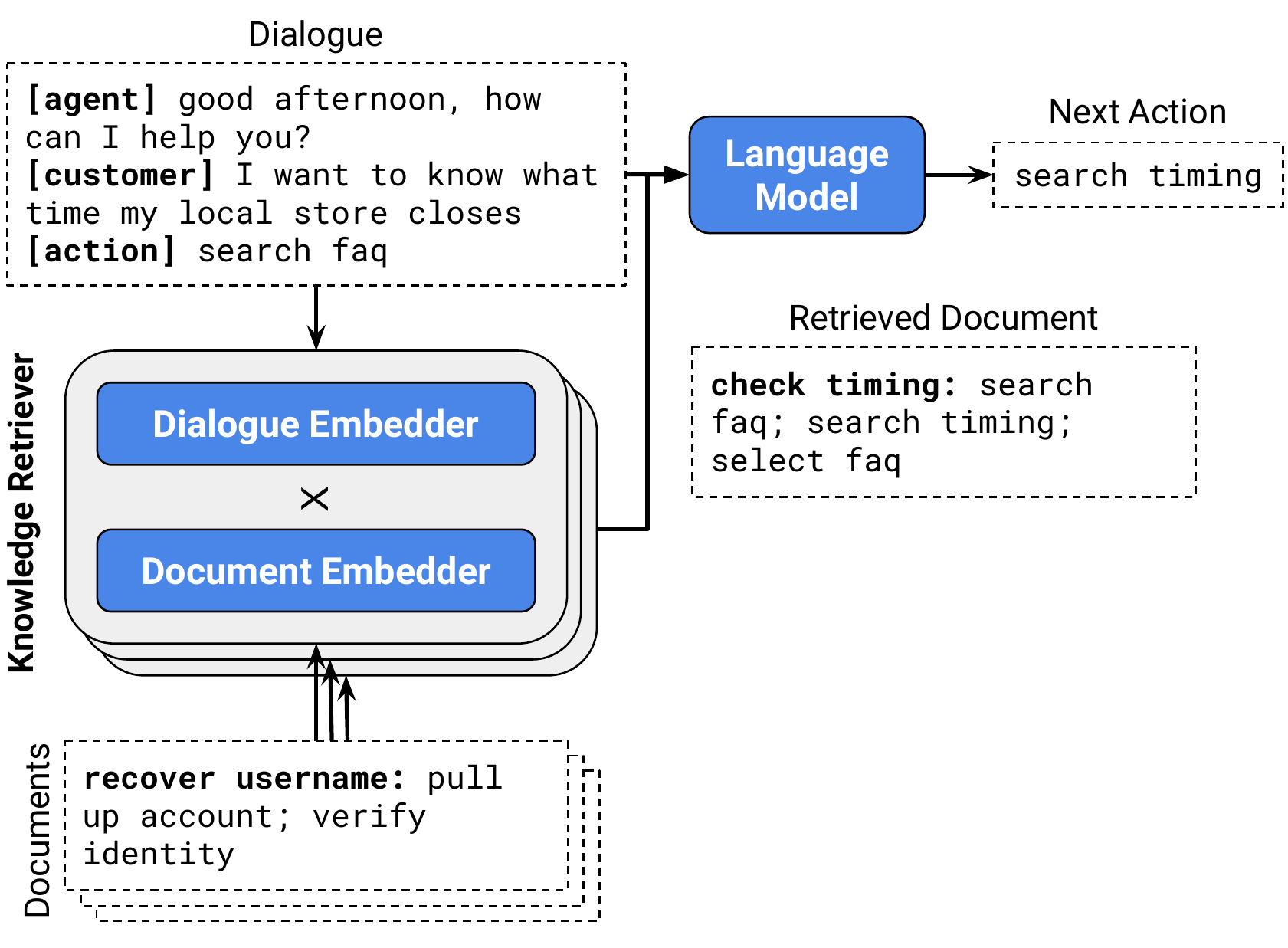}
\end{center}
\caption{The Knowledge-Augmented Dialogue System (KADS) is composed of two modules: a knowledge retriever and a language model. The knowledge retriever takes the inner product as a measure of similarity between an embedded dialogue and each document in a provided knowledge base containing procedural instructions. The most similar document is then passed to a language model which attends over both the dialogue and retrieved document to generate the agent's next action.}
\label{arch}
\end{figure} 

\section{Introduction}

For many real-world applications, it is crucial for task-oriented dialogue (TOD) systems to complete user requests while strictly adhering to established procedures. For example, consider a customer service agent who must first verify a client's details before changing their password. Although large language models have demonstrated potential in modeling such dialogues, they require large amounts of data with consistent procedural representations to \textit{implicitly} store procedures in the parameters of their underlying networks. In practical settings, such high-quality data is not always readily available as some procedures may naturally occur infrequently or change over time. In this paper, we explore a solution to TOD modeling which improves performance in low-data settings by referencing \textit{explicitly} stored agent guidelines.

We outline a methodology of incorporating procedural knowledge (i.e., knowledge concerning the requisite steps to address a user inquiry) into a language model with the objective of predicting agent actions in dialogue tasks. Our proposed system, the Knowledge-Augmented Dialogue System (KADS), consists of two modules: a knowledge retriever which, given a dialogue between an agent and user, retrieves the most pertinent instructions from a knowledge base of agent procedures and a language model which considers the retrieved instructions along with the ongoing dialogue to inform an action prediction (see architecture in \autoref{arch}). 

In prior work, retrieval-enhanced language models have achieved success integrating external knowledge from internet searches into conversational agents \cite{shuster2022,thoppilan2022}. However, a more controllable approach is necessary for instruction retrieval in task-oriented dialogue. Rather than querying the open web, it's more suitable to perform retrieval over a closed set of documents, like in \cite{guu2020,lewis2020}. However, while the training schemes utilized in these works sufficiently prime a model for question-answering tasks, they are not as effective for action prediction. 

Following the lines of \cite{henderson2021}, which introduces a unique pre-training objective for slot-labeling, our method leverages custom objectives suited for action prediction tasks. We employ a specialized warm-up task where dialogues are matched with corresponding procedural instructions to ensure that the knowledge retrieval module is initialized with reasonable dialogue and document embeddings. Then, the system is trained on an special case of masked language modeling in which masked actions are predicted from customer-agent dialogues. Finally, we found it necessary to encourage our system to incorporate signal from retrieved procedures by routinely freezing the language model's weights during training.

We evaluated this approach on two dialogue tasks--- action state tracking and workflow discovery--- using two task-oriented dialogue datasets: Action-Based Conversations Dataset and Schema-Guided Dialogue. Our results suggest that KADS yields improved action prediction accuracy against several baselines, including an un-augmented language model and a language model augmented with static guidelines, on both in- and out-of-distribution procedures. Furthermore, we demonstrate that knowledge augmentation bolsters our system's ability to predict actions that occur infrequently in the training data.

\section{Dialogue Tasks}

TOD systems are employed for a variety of tasks including action state tracking and workflow discovery.

\textbf{Action state tracking (AST)} aims to predict the next action performed by an agent during an interaction with a customer \cite{chen2021}. Formally, we represent an interaction as a sequence of turns $x$ belonging to one of three categories: agent utterances $x^a$ (\texttt{[agent]}), agent actions $x^b$ (\texttt{[action]}), or customer utterances $x^c$ (\texttt{[customer]}).
The model receives an interaction between a customer and agent up to turn $t$ where prefix tokens $p$ indicate the turn category: $X = p_0 \; x_0 \; p_1 \; x_1 \; ... \; p_t \; x_t$ with $p \in {\texttt{[agent]}, \texttt{[action]}, \texttt{[customer]}}$. The model then predicts the following agent action $x^b_{t+1}$ which consists of a button, or b-slot, and any corresponding slot values if they are present: $x^b_t = b_t^0: v_t^{00}, v_t^{01}$.

The goal of \textbf{workflow discovery (WD)} is to recover the workflow--- the set of ordered actions taken by an agent--- given a complete dialogue between a customer and agent \cite{hattami2022}. 
Formally, we represent a dialogue as a sequence of turns belonging to one of two categories: agent utterances or customer utterances.
The model receives a dialogue of length $T$ between a customer and agent where prefix tokens indicate the turn category: $X = p_0 \; x_0 \; p_1 \; x_1 \; ... \; p_T \; x_T$ with $p \in {\texttt{[agent]}, \texttt{[customer]}}$. 
The model then predicts the corresponding agent actions $x^b_{0}; x^b_{1}; ...; x^b_{T}$.

\section{Approach}

\subsection{Architecture} 
The end goal of KADS is to learn a distribution $p(y|X)$ over possible action sequences $y$ given an interaction or dialogue $X$.
Our approach utilizes a knowledge retriever module to produce a relevance score between a given procedural document $z$ and $X$. We calculate the relevance score according to \cite{devlin2019} as the inner product of the BERT vector embeddings of $X$ and $z$. A retrieval distribution $p(z|X)$ is obtained by taking the softmax over the relevance scores corresponding to each available document and the given interaction or dialogue.
Finally, we train a T5 language model \cite{raffel2020}, conditioned on both the retrieved document $z$ and the interaction $X$, to generate an action sequence $y$, where the likelihood of generating $y$ is obtained by treating $z$ as a latent variable and marginalizing over all possible documents: $p(y|X)=\sum_{z\in Z}p(y|X,z)p(z|X)$.

\subsection{Training} 
To train KADS we follow a three-step procedure: first, we warm-up the knowledge retriever's embedding modules with a dialogue-document matching task; then, we pre-train the full model with action-oriented masked language modeling (MLM); finally, we train on one of two downstream dialogue tasks--- AST or WD. For all tasks except dialogue-document matching, our training objective is to maximize the log-likelihood $\text{log}p(y|X)$ of the correct output action sequence $y$. However, calculating the marginal probability over documents in a knowledge corpus can become costly as the number of documents grows, so we approximate this probability by summing over the top $5$ documents with the highest probability under $p(z|X)$. We then compute the gradient of the log-likelihood with respect to the model parameters of both the knowledge retriever and language model and optimize using stochastic gradient descent.

We first perform the dialogue-document matching warm-up routine to ensure that the knowledge retriever is initialized with reasonable dialogue and document embeddings. The embedding modules are pre-trained using a semi-supervised training procedure with the objective of retrieving the document that most likely corresponds to a specific dialogue. This label is determined according to which document has the highest action overlap with the dialogue or, when provided, which document corresponds to the user's ground-truth intent. 

For the MLM pre-training task, we randomly mask action sequences from dialogue transcripts such that the system learns to retrieve relevant documents in order to better predict the actions corresponding to each \texttt{[MASK]} token. To prevent KADS from learning to ignore retrieved documents we employ several tricks during MLM training. First, we filter out dialogues with action sequences that are not detailed in the agent guidelines. This is done to ensure that only examples in which the knowledge retriever may be useful are present.
Additionally, we freeze the language model weights with $0.9$ probability to encourage updates to the knowledge retriever parameters which minimize the MLM loss. 

\section{Data}
We evaluate KADS on two TOD datasets: Action-Based Conversations Dataset and Schema-Guided Dialogue. 
Both consist of multi-domain customer service interactions that loosely follow a set of predefined company policies which specify the actions to be taken by an agent to satisfy a particular customer inquiry. The core differences between these two datasets are their action and document structures.

In \textbf{Action-Based Conversations Dataset (ABCD)} \cite{chen2021}, actions are composed such that the b-slot belongs to a predefined set of b-slots which describe the action being taken (e.g., "pull up account") and slot values consist of any corresponding information provided by the user (e.g., "johndoe@gmail.com"). In a given interaction, an average of $4$ actions are taken. The documents provided within ABCD are composed of a plain text description of a possible customer inquiry followed by an ordered set of action b-slots that should be performed by the agent.

In \textbf{Schema-Guided Dialogue (SGD)} \cite{rastogi2020}, we take action b-slots to be the description of how the agent will interact with a piece of information (e.g., "inform", "confirm", or "request") and values as the type of information in question (e.g., "departure times"). In this dataset, the average number of actions per interaction is significantly longer at $21$ actions, and the documents corresponding to SGD consist of a customer inquiry followed by all of the information types, or values, that can be acquired to fulfill the given inquiry. 

We use the train/dev/test splits presented in the original datasets ($8034/1004/1004$ and $16142/2482/4201$ interactions per split for ABCD and SGD respectively), and hold out a randomly-selected subset of $10\%$ of actions during training for out-of-distribution testing. See \autoref{sec:data} for more details, including dialogue and corresponding document examples.

\begin{table*}[t]
\begin{center} 
\caption{B-Slot and value prediction accuracy on in-distribution actions.} 
\label{id}
\vskip 0.12in
\begin{tabular}{l|cc|cc||cc|cc} 
\hline
\multicolumn{1}{c|}{\multirow{3}{*}{\textbf{Model}}} & \multicolumn{4}{c||}{\textbf{AST}} & \multicolumn{4}{c}{\textbf{WD}}\\
\cline{2-9}
& \multicolumn{2}{c|}{\textbf{ABCD}} &
\multicolumn{2}{c||}{\textbf{SGD}} & 
\multicolumn{2}{c|}{\textbf{ABCD}} & 
\multicolumn{2}{c}{\textbf{SGD}}\\
\cline{2-9}
& \textbf{B-Slot} & \textbf{Value} & \textbf{B-Slot} & \textbf{Value} & \textbf{B-Slot} & \textbf{Value} & \textbf{B-Slot} & \textbf{Value} \\ 
\hline
\hline
T5 & 79.5 & 82.2 & 51.8 & 31.6 & 65.9 & 66.8 & \textbf{58.7} & \textbf{28.3} \\
T5 + guide & 81.3 & 82.5 & NA & NA & 56.8 & 58.4 & NA & NA \\
KADS & \textbf{85.2} & \textbf{83.1} & \textbf{63.2} & \textbf{39.5} & \textbf{72.5} & \textbf{73.0} & 53.1 & 23.8 \\
\hline
\end{tabular}
\end{center} 
\end{table*}

\section{Results}

The evaluation of our TOD system begins with b-slot and value prediction accuracy for both known and novel actions.
We also examine the data efficiency of our approach by reporting these metrics for progressively reduced training pools.
We compare our model's performance against a base T5 model and T5 with static guidelines--- a comprehensive list of agent actions--- appended to the input sequence (T5 + guide)\footnote{Appending static guidelines is not possible for SGD, where the potential action space is too large to fit within the maximum input sequence length.}. Then, we assess the efficacy of our knowledge retriever in selecting relevant documents. Finally, an ablation study of our pre-training routine highlights the importance of our custom training procedure.
See \autoref{sec:experiment} for details of our experimental setup.

\subsection{In-Distribution Performance}
We first observe b-slot and value prediction accuracy on procedures observed during training (\autoref{id}).

On ABCD, KADS achieves higher b-slot prediction accuracy than our baselines for both tasks. The inclusion of a static guideline offers slightly improved accuracy on AST but is not nearly as effective as the dynamic guide provided by the knowledge retriever.
We attribute the performance boost in part to KADS's ability to predict actions that are less represented during training. 

\begin{figure}[t]
\begin{center}
\includegraphics[width=7.5cm]{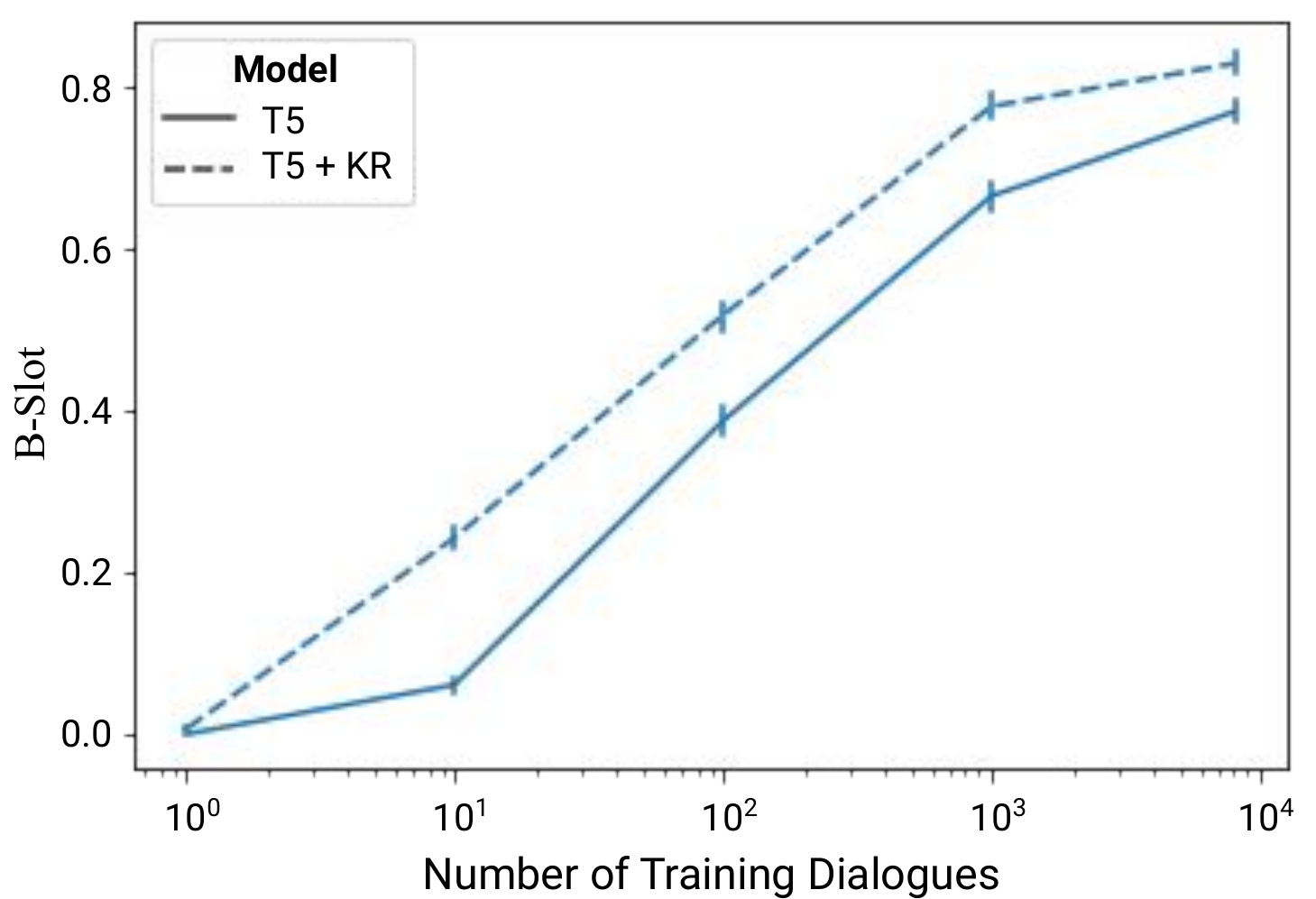}
\end{center}
\caption{B-Slot accuracy on AST task trained with varying numbers of ABCD dialogues. Error bars represent 95\% confidence interval.}
\label{dataeff}
\end{figure}

This characteristic is evidenced by the model's performance in low-data settings (\autoref{dataeff}). We observe that the difference in action prediction accuracy between our model and the unaugmented baseline increases when training on progressively fewer dialogues. Additionally, we find that, for the base and static guide models, the correlation between a b-slot's level of occurrence in the training data and the model's accuracy in predicting that b-slot is notably higher ($0.27$ and $0.24$ respectively) than in the knowledge-augmented model ($0.18$). We conclude from these results that KADS is more robust to low-data settings where the quantity of individual action occurrences is low or inconsistent.  \footnote{Value prediction accuracy is improved despite values not being included in the provided documents. This is likely a result of the model learning patterns between action b-slots and their corresponding values.}

On SGD, we see similar trends for the AST task. However, for the WD task, which concerns recovering the entire action sequence from a dialogue at once, we see that knowledge augmentation does not provide substantial improvement in performance. This may be due to the nature of SGD dialogues, which contain multiple client requests, while the model is augmented with a singular document providing instructions for a singular customer request. 

\subsection{Out-of-Distribution Performance}
Next, we evaluate the ability of KADS to generalize to novel procedures by assessing performance on actions not seen during training (\autoref{ood}). Both tasks, AST and WD, show knowledge augmentation to improve novel b-slot prediction accuracy over the baselines, coming only second to T5 trained on the full dataset (“full data”) including “out-of-distribution” actions. These results demonstrate that KADS is able to relatively accurately predict new actions in a zero-shot fashion by making use of documents containing information about the action.

\begin{table}[t]
\begin{center} 
\caption{B-Slot and value prediction accuracy on out-of-distribution actions.} 
\label{ood}
\vskip 0.12in
\begin{tabular}{l|cc|cc} 
\hline
\multicolumn{1}{c|}{\multirow{2}{*}{\textbf{Model}}} & \multicolumn{2}{c|}{\textbf{ABCD}} &
\multicolumn{2}{c}{\textbf{SGD}} \\
\cline{2-5}
& \textbf{B-Slot} & \textbf{Value} & \textbf{B-Slot} & \textbf{Value} \\ 
\hline
\hline
T5 & ~0.0 & 11.6 & 46.2 & 25.8 \\
T5 + guide & ~2.7 & 17.9 & NA & NA \\
KADS & \textbf{11.6} & \textbf{21.4} & \textbf{49.8} & \textbf{32.2} \\
\hline
full data & 94.6 & 85.7 & 61.3 & 38.1 \\
\hline
\end{tabular}
\end{center} 
\end{table}

\subsection{Document Selection Accuracy}
We use document selection accuracy to assess how well our knowledge retriever selects documents that correspond to a customer's inquiry. On ABCD, we define the correct document as the document with the most action b-slots overlapping with the full customer-agent interaction. On SGD, where calls often consist of multiple customer inquiries, the correct document is instead defined as the document corresponding to the labeled customer intent for any given step of the interaction. In \autoref{docacc}, we see that approximate document selection accuracy on the AST task for ABCD is significantly higher than that of SGD. This is likely due to the significant overlap in procedures for similar customer inquiries on the latter dataset. For example, making an appointment with a doctor, dentist, or hairstylist requires similar values to be filled, which results in related documents being somewhat interchangeable for these inquiries.

Furthermore, we measure document selection accuracy on our pre-training tasks: dialogue-document matching and MLM. Notably, the knowledge retriever's document selection accuracy decreases between pre-training with the dialogue-document matching task and fine-tuning on the final task. This is likely due to the objective changing from maximizing document selection accuracy to predicting correct action sequences, resulting in some drift from the selection of approximated "correct" documents.

\begin{table}[t]
\begin{center} 
\caption{Document retrieval accuracy for the ABCD and SGD datasets after training with dialogue-document matching (DDM), MLM, and AST.} 
\label{docacc}
\vskip 0.12in
\begin{tabular}{l|ccc} 
\hline
\multirow{2}{*}{\textbf{Dataset}} & \multicolumn{3}{c}{\textbf{Accuracy}} \\
\cline{2-4}
& \textbf{DDM} & \textbf{MLM} & \textbf{AST} \\
\hline
\hline
\textbf{ABCD} & 98.0 & 82.3 & 85.4 \\
\textbf{SGD} & 66.2 & 74.9 & 63.9 \\ 
\hline
\end{tabular}
\end{center} 
\end{table}

\subsection{Pre-training Scheme Ablations}

Our full training scheme is a multi-step process ensuring optimal performance from our Knowledge-Augmented Dialogue System. First, the knowledge retrieval module is tuned on a dialogue-document matching task to ensure that the model is initialized with sensible dialogue and document embeddings. Next, the full system is trained on an MLM task which acts as the simpler in-between before our final task. Finally, we train the model for one of our two downstream dialogue tasks.
Removing any step from this procedure results in decreased performance on the final task. 
In \autoref{pretraining}, we share b-slot and value prediction accuracy on AST after pre-training with several ablations of our full scheme. These results show that the elimination of either the dialogue-document matching or MLM task results in lower accuracy. These tasks, which allow our model to effectively harness the knowledge retrieval module, are crucial to our pre-training procedure.

\begin{table}[t]
\begin{center} 
\caption{AST task b-slot and value prediction accuracy for the ABCD dataset after training with several ablations of our pre-training scheme.} 
\label{pretraining}
\vskip 0.12in
\begin{tabular}{l|cc} 
\hline
\textbf{Model} & \textbf{B-Slot} & \textbf{Value} \\
\hline
\hline
none & 82.7 & 79.4 \\
MLM only & 81.5 & 79.0\\
DDM only & 82.6 & 78.5 \\
\textbf{full} & \textbf{85.2} & \textbf{83.1} \\
\hline
\end{tabular}
\end{center} 
\end{table}

\section{Conclusion}
While large language models make for effective TOD systems in constrained settings, real-world applications often present insufficient data to train these models.
KADS offers a method of learning workflows with minimal or sparse supporting data and presents a more controllable and performant solution to low-resource TOD automation. 
While our results offer a promising outlook for action prediction given dynamic guidance from structured procedural documents, future work should investigate the use of unstructured company guidelines and multi-document retrieval.

\section{Limitations}
Our paper assesses procedural knowledge augmentation using a limited number of highly structured instructional documents. Naturally, the results presented may vary for unstructured guidelines. Additionally, due to the limited size of publicly available TOD datasets, we have not tested how our method may scale to settings with larger document spaces ($>100$ documents). For larger document sets, more efficient methods of computing similarity such as Maximum Inner Product Search (MIPS) algorithms may be necessary to approximate documents with the highest relevance scores.

\bibliography{anthology,custom}
\bibliographystyle{acl_natbib}

\newpage

\appendix

\begin{table*}
\begin{center} 
\caption{Example AST input and output sequences.} 
\label{astex}
\vskip 0.12in
\begin{tabular}{l|l} 
\multicolumn{2}{c}{\textbf{ABCD}}  \\
\hline
\textbf{Input} & [agent] hello! how can i help you today? [customer] i'm thinking about buying an item \\
&  but first i would like to get some more info on the product [agent] sure. i can help you \\
&  with that. what item are you looking for more information on? [customer] the tommy \\
& hilifiger shirt [agent] and what would you like to know about it? [customer] i would \\
& like to know how long is the arm length [agent] sure give me one second and i can find \\
& that out for you [customer] ok [action] search faq \\
\hline
\textbf{Output} & search shirt \\
\hline
\textbf{Document} & get shirt info [SEP] search faq; search shirt; select faq \\
\hline
\multicolumn{2}{c}{}  \\
\multicolumn{2}{c}{\textbf{SGD}}  \\
\hline
\textbf{Input} & [customer] i am interested to know how the weather is going to be on 7th of march in \\
& san diego. \\
\hline
\textbf{Output} & offer temperature; offer precipitation \\
\hline
\textbf{Document} & get the weather of a certain location on a date [SEP] [required] city [optional] date \\
& [result] precipitation; humidity; wind; temperature; city; date \\
\hline
\end{tabular}
\end{center} 
\end{table*}

\section{Experimental Details}
\label{sec:experiment}
Our implementations are based on the Hugging Face Transformer models \cite{huggingface}. Each embedding module in the knowledge retriever is a small BERT model with 4 layers and a hidden size of 512, and the language model used is a pre-trained T5 model, \textit{t5-base}. All models were trained with a learning rate of $0.00001$ using the AdamW optimizer and an effective batch size of $32$. We used an NVIDIA TITAN X GPU for all experiments. 

\section{Data Details}
\label{sec:data}

We evaluate on two TOD datasets: Action-Based Conversations Dataset (ABCD) and Schema-Guided Dialogue (SGD)--- each with a slightly different composition. 

ABCD contains over 10,000 human-to-human customer service dialogues across multiple domains. The agent's actions are constrained to a set of 30 action b-slots and unrestricted, free-form slot values. There are a total of 55 structured documents relating recommended sequences of action b-slots to various customer inquiries.

SGD contains over 20,000 multi-domain conversations between a human and a virtual assistant. There are 8 possible action b-slots and 132 possible slot values. There are a total of 53 documents containing the required and optional slot values to collect in order to fulfill a specific customer intent.

\begin{figure}[h!]
\begin{center}
\includegraphics[width=7.5cm]{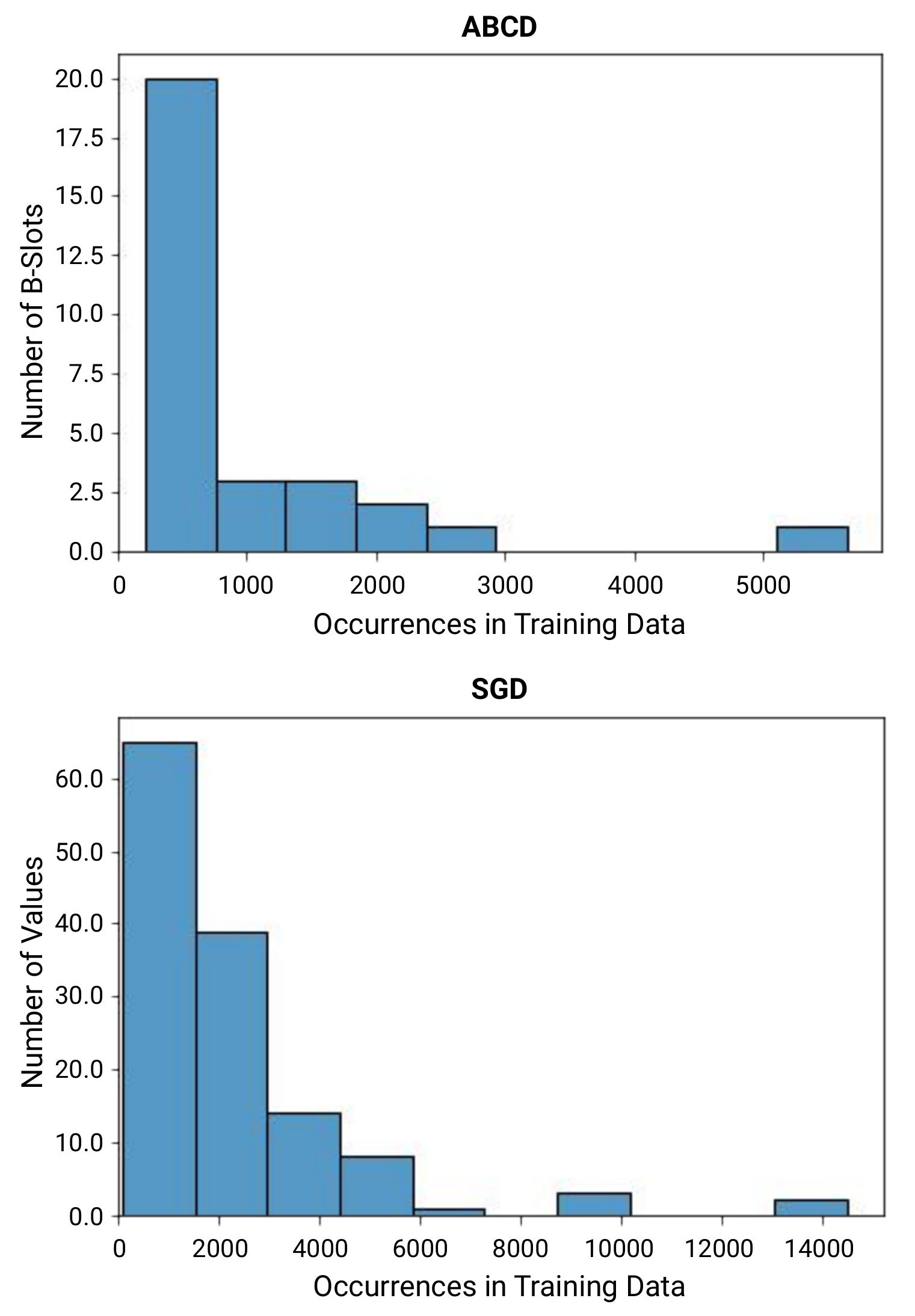}
\end{center}
\caption{Distribution of action B-slots and values for the ABCD and SGD datasets.}
\label{actdist}
\end{figure} 

Example AST input and output sequences for both datasets are provided in \autoref{astex}: these include the input interaction between a customer and agent, the output next agent action, and the corresponding document. The distribution of actions (b-slots and slot values for ABCD and SGD respectively) indicate an imbalance in both datasets with some actions being significantly more represented than others (\autoref{actdist}).



\end{document}